\documentclass[runningheads]{llncs}
\usepackage{graphicx}
\usepackage{amsmath,amssymb} 
\usepackage{color} 
\usepackage{hyperref}

\usepackage[width=122mm,left=12mm,paperwidth=146mm,height=193mm,top=12mm,paperheight=217mm]{geometry}

\usepackage{multirow}
\usepackage{array}
\newcolumntype{P}[1]{>{\centering\arraybackslash}p{#1}}

\begin{document}
\title{Conditional Prior Networks for Optical Flow}
\titlerunning{CPNFlow}
\author{Yanchao Yang \ \ \ Stefano Soatto}
\authorrunning{Y. Yang and S. Soatto}
\institute{UCLA Vision Lab\\ University of California, Los Angeles, CA 90095\\
\email{\{yanchao.yang, soatto\}@cs.ucla.edu}}
\maketitle

\begin{abstract}
Classical computation of optical flow involves generic priors (regularizers) that capture rudimentary statistics of images, but not long-range correlations or semantics. On the other hand, fully supervised methods learn the regularity in the annotated data, without explicit regularization and with the risk of overfitting. We seek to learn richer priors on the set of possible flows that are statistically compatible with {\em an} image. Once the prior is learned in a supervised fashion, one can easily learn the full map to infer optical flow directly from {\em two or more} images, without any need for (additional) supervision. We introduce a novel architecture, called Conditional Prior Network (CPN), and show how to train it to yield a conditional prior. When used in conjunction with a simple optical flow architecture, the CPN beats all variational methods and all unsupervised learning-based ones using the same data term. It performs comparably to fully supervised ones, that however are fine-tuned to a particular dataset. Our method, on the other hand, performs well even when transferred between datasets. Code is available at: \url{https://github.com/YanchaoYang/Conditional-Prior-Networks}
\end{abstract}

\begin{figure}[!ht]
  \centering
  \includegraphics[width=0.85\textwidth]{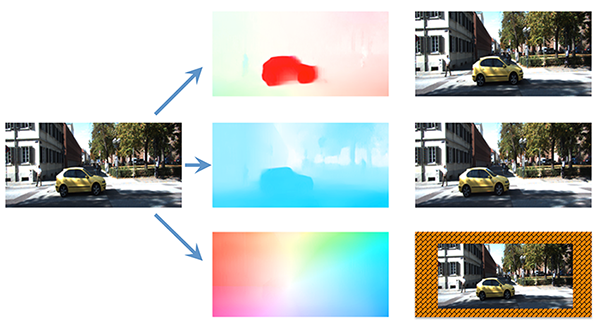}
\caption{A single image biases, but does not constrain, the set of optical flows that can be generated from it, depending on whether the camera was static but objects were moving (top), or the camera was moving (center), or the scene was flat (bottom) and moving on a plane in an un-natural scenario. Flow fields here are generated by our CPNFlow.}
\label{fig-1}
\end{figure}

\section{Introduction}

Consider Fig. \ref{fig-1}: A given image (left) could give rise to many different optical flows (OF) \cite{horn1981determining} depending on what another image of the same scene looks like: It could show a car moving to the right (top), or the same apparently moving to the left due to camera motion to the right (middle), or it could be an artificial motion because the scene was a picture portraying the car, rather than the actual physical scene. A single image biases, but does not constrain, the set of possible flows the underlying scene can generate. We wish to leverage the information an image contains about possible compatible flows to learn better priors than those implied by generic regularizers. Note that all three flows in Fig. \ref{fig-1} are equally valid under a generic prior (piecewise smoothness), but not under a natural prior (cars moving in the scene).

A regularizer is a criterion that, when added to a data fitting term, constrains the solution of an inverse problem. These two criteria (data term and regularizer) are usually formalized as an energy function, which is minimized to, ideally, find a unique global optimum.\footnote{We use the terms regularizer, prior, model, or assumption, interchangeably and broadly to include any restriction on the solution space, or bias on the solution, imposed without full knowledge of the data. In OF, the full data is (at least) two images.}

\subsection{Our approach in context}

In classical (variational) OF, the regularizer captures very rudimentary low-order statistics \cite{baker2011database,bruhn2005lucas,papenberg2006highly,black1993framework,xu2012motion}, for instance the high kurtosis of the gradient distribution. This does not help with the scenario in Fig. \ref{fig-1}. There has been a recent surge of (supervised) learning-based approaches to OF \cite{dosovitskiy2015flownet,ilg2017flownet,ranjan2017optical}, that do not have explicit regularization nor do they use geometric reprojection error as a criterion for data fit. Instead, a map is learned from pairs of images to flows, where regularization is implicit in the function class \cite{cohen2016inductive},\footnote{In theory, deep neural networks are universal approximants, but there is a considerable amount of engineering in the architectures to capture suitable inductive biases.} in the training procedure \cite{chaudhari2017stochastic} (e.g. noise of stochastic gradient descent -- SGD), and in the datasets used for training (e.g. Sintel \cite{butler2012naturalistic}, Flying Chair \cite{dosovitskiy2015flownet}).

Our method does not attempt to learn geometric optics anew, even though black-box approaches are the top performers in several benchmarks. Instead, we seek to learn richer priors on the set of possible flows that are statistically compatible with an image (Fig. \ref{fig-1}).

Unsupervised learning-based approaches use the same or similar loss functions as variational methods \cite{jason2016back,ren2017unsupervised,Meister:2018:UUL,ahmadi2016unsupervised}, including priors, but restrict the function class to a parametric model, for instance convolutional neural networks (CNNs) trained with SGD, thus adding implicit regularization \cite{chaudhari2017stochastic}. Again, the priors only encode first-order statistics, which fail to capture the phenomena in Fig. \ref{fig-1}.

We advocate learning a conditional prior, or regularizer, from data, but do so once and forall, and then use it in conjunction with any data fitting term, with any model and optimization one wishes.

What we learn is a prior in the sense that it imposes a bias on the possible solutions, but it does not alone constraint them, which happens only in conjunction with a data term. Once the prior is learned, in a supervised fashion, one can also learn the full map to infer optical flow directly from data, without any need for (additional) supervision. In this sense, our method is {\em ``semi-unsupervised''}: Once {\em we} learn the prior, {\em anyone} can train an optical flow architecture entirely unsupervised. The key idea here is to learn a prior for the set of optical flows that are statistically compatible with {\em a single image}. Once done, we train a relatively simple network {\em in an unsupervised fashion} to map {\em pairs of images} to optical flows, where the loss function used for training includes explicit regularization in the form of the conditional prior, added to the reprojection error.

Despite a relatively simple architecture and low computational complexity, our method beats all variational ones and all unsupervised learning-based ones. It is on par or slightly below a few fully supervised ones, that however are fine-tuned to a particular dataset, and are extremely onerous to train. More importantly, available fully supervised methods perform best {\em on the dataset on which they are trained.} Our method, on the other hand, performs well even when the prior is trained on one dataset and used on a different one. For instance, a fully-supervised method trained on Flying Chair beats our method on Flying Chair, but underperforms it on KITTI and vice-versa (Tab. \ref{tab:quantative-comparison-all}). Ours is consistently among the top in all datasets. More importantly, our method is complementary, and can be used in conjunction with more sophisticated networks and data terms.

\subsection{Formalization}

\begin{figure}[!ht]
  \centering
  \includegraphics[width=0.7\textwidth]{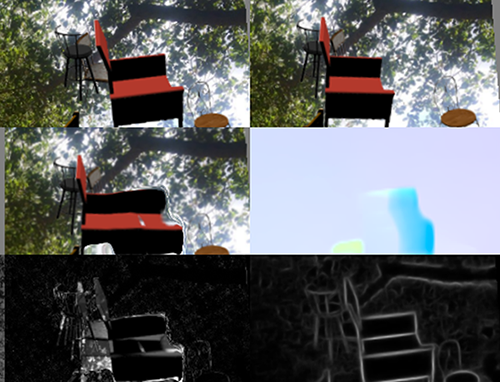}
  \caption{First row: two images $I_1, I_2$ from the Flying Chair dataset; Second row: warped image $I_2 \circ \hat f$ (left) using the flow (right) estimated by minimizing Eq. \eqref{eq:classical-of}; Third row: residual $n = \| I_1 - I_2 \circ f \|$ (left) compared to the edge strength of $I_1$ (right). Note the flow estimated at the right side of the chair fails to propagate into the homogeneous region where the image gradient is close to zero.}
  \label{fig:drawback-smoothness}
\end{figure}

Let $I_1, I_2 \in \mathbb{R}_+^{H \times W \times 3}$ be two consecutive images and $f : \mathbb{R}^2 \rightarrow \mathbb{R}^2$ the flow, implicitly defined in the co-visible region by $I_1 = I_2 \circ f + n$  where $n \sim P_n$ is some distribution. The posterior $P(f|I_1, I_2) \propto P_n(I_1 - I_2 \circ f)$ can be decomposed as 
\begin{multline}
\log P(f|I_1, I_2) = \log P(I_2|I_1, f) + \log P(f|I_1) - \log P(I_2|I_1)\\
\approx \log P(I_2|I_1, f) + \log P(f|I_1)
\label{eq:posterior}
\end{multline}
We call the first term (data) \textbf{prediction error}, and the second \textbf{conditional prior}. It is a prior in the sense that, given $I_1$ alone, many flows can have high likelihood for a suitable $I_2$. However, it is informed by $I_1$ in the sense of capturing image-dependent regularities such as flow discontinuities often occurring at {\em object boundaries}, which may or may not correspond to generic image discontinuities. A special case of this model assumes a Gaussian likelihood ($\ell^2$ prediction error) and an ad-hoc prior of the form 
\begin{multline}
E(f, I_1, I_2)=\int ( I_1(x) - I_2(x+f(x)) )^2dx + \int \alpha(x, I_1) \lVert \nabla f(x) \rVert^2 dx
\label{eq:classical-of}
\end{multline}
where $\alpha$ is a scalar function that incorporates our belief in an irradiance boundary of $I_1$ corresponding to an object boundary.\footnote{When $\alpha$ is constant, we get an even more special case, the original Horn \& Schunk model where the prior is also Gaussian and unconditional (independent of $I_1$).} This type of conditional prior has several limitations: First, in the absence of {\em semantic context}, it is not possible to differentiate occluding boundaries (where $f$ can be discontinuous) from material boundaries (irradiance discontinuities), or illumination boundaries (cast shadows) where $f$ is smooth. Second, the image $I_1$ only informs the flow {\em locally}, through its gradient, and does not capture global regularities. Fig. \ref{fig:drawback-smoothness} shows that flow fails to propagate into homogeneous region. This can be mitigated by using a fully connected CRF \cite{sutton2012introduction} but at a heavy computational cost.

Our goal can be formalized as {\em learning the conditional prior} $P(f|I_1)$ in a manner that exploits the semantic context of the scene\footnote{The word ``semantic'' is often used to refer to {\em identities} and {\em relations} among discrete entities (objects). What matters in our case is the {\em geometric and topological relations} that may result in occluding boundaries on the image plane. The name of an object does not matter to that end, so we ignore identities and do not require object labels.} and captures the global statistics of $I_1$. We will do so by leveraging the power of deep convolutional neural networks trained end-to-end, to enable which we need to design differentiable models, which we do next.

\section{Method}

To learn a conditional prior we need to specify the inference criterion (loss function), which we do in Sect. \ref{sect-loss} and the class of functions (architecture), with respect to which the loss is minimized end-to-end. We introduce our choice of architecture next, and the optimization in Sect. \ref{sect-training}.

\subsection{Conditional Prior Network (CPN)}

\begin{figure}[!ht]
  \centering
  \includegraphics[width=0.42\textwidth]{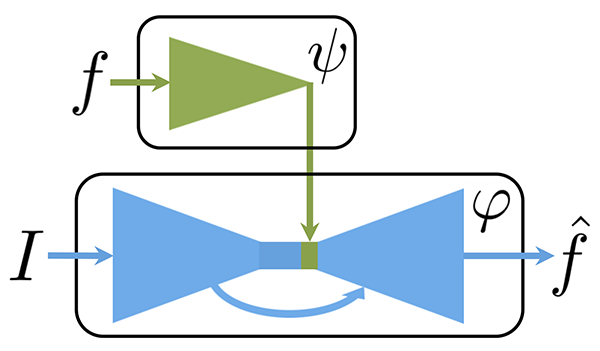}
  \caption{Conditional Prior Network (CPN) architecture for learning $P(f|I)$: $\psi$ is an encoder of the flow $f$, and $\varphi$ is a decoder that has full access to the image $I$.}
  \label{fig:CPNet}
\end{figure}

We construct the conditional prior from a modified autoencoder trained to reconstruct a flow $f$ that is compatible with the given (single) image $I$. We call this a Conditional Prior Network (CPN) shown in Fig. \ref{fig:CPNet}.

In a CPN, $\psi$ encodes only the flow $f$, then $\varphi$ takes the image $I$ and the output of $\psi$ to generate a reconstruction of $f$,  $\hat{f} = \varphi( I, \psi(f))$. Both $\psi$ and $\varphi$ are realized by pure convolutional layers with subsampling (striding) by two to create a bottleneck. Note that $\varphi$ is a U-shape net \cite{dosovitskiy2015flownet} with skip connections, at whose center a concatenation with $\psi(f)$ is applied. In the appendix, we articulate the reasons for our choice of architecture, and argue that it is better than an ordinary autoencoder that encodes both $f$ and $I$ in one branch. This is connected to the choice of loss function and how it is trained, which we discuss next. 

\subsection{Loss function}
\label{sect-loss}

We are given a dataset $D$ sampled from the joint distribution $D=\{ (f_j, I_j) \}_{j=1}^n \sim P(f, I)$, with $n$ samples. We propose approximating $P(f|I)$ with a CPN as follows 
\begin{equation}
Q_{w_{\varphi},w_{\psi}}(f|I) = \exp \left( - \| \varphi(I, \psi(f)) - f \|^2 \right) \propto P(f|I)
\label{eq:proposed-conditional-prior}
\end{equation}
where $w_{\varphi},w_{\psi}$ are the parameters of $\varphi$ and $\psi$ respectively. Given $I$, for every flow $f$, the above returns a positive value whose log, after training, is equal to the negative squared autoencoding loss. To determine the parameters that yield an approximation of $P(f|I)$, we should solve the following optimization problem
\begin{equation}
w_{\varphi}^*, w_{\psi}^* = \arg\min_{w_{\varphi}, w_{\psi}} {\mathbb E}_{I\sim P(I)} \mathbb{KL}( P(f|I) \| Q_{w_{\varphi}, w_{\psi}}(f|I) )
\label{eq:optimization-CPN}
\end{equation}
where the expectation is with respect to all possible images $I$, and $\mathbb{KL}$ is the Kullback-Leibler divergence between $P(f|I)$ and the CPN $Q_{w_{\varphi}, w_{\psi}}(f|I)$. In the appendix we show that the above is equivalent to:
\begin{multline}
w_{\varphi}^*, w_{\psi}^* = \arg\max_{w_{\varphi}, w_{\psi}} \int_I \int_f P(f, I) \log [ Q_{w_{\varphi}, w_{\psi}}(f|I) ]df dI \\
= \arg\min_{w_{\varphi}, w_{\psi}} \int_I \int_f P(f, I) \| \varphi_{w_\varphi}(I, \psi_{w_\psi}(f)) - f \|^2 df dI
\label{eq:equivalent-optimization-CPN}
\end{multline}
which is equivalent to minimizing the empirical autoencoding loss since the ground truth flow is quantized, $\sum_{j=1}^n \| \hat{f}_j - f_j \|^2$. If the encoder had no bottleneck (sufficient information capacity), it could overfit by returning $\hat{f} = \varphi_{w_\varphi}(I, \psi_{w_\psi}(f)) = f$, rendering the conditional prior $Q_{w_{\varphi},w_{\psi}}(f|I)$ uninformative (constant). Consistent with recent developments in the theory of Deep Learning \cite{achille2017emergence}, sketched in the appendix, we introduce an information regularizer (bottleneck) on the encoder $\psi$ leading to the \textbf{CPN training loss}
\begin{equation}
w_{\varphi}^*, w_{\psi}^* = \arg\min_{w_{\varphi}, w_{\psi}} {\mathbb E}_{I\sim P(I)} \mathbb{KL}( P(f|I) \| Q_{w_{\varphi}, w_{\psi}}(f|I) ) + \beta \boldsymbol{I}(f,\psi_{w_{\psi}}(f))
\label{eq:final-optimization-CPN}
\end{equation}
where $\beta > 0$ modulates complexity (information capacity) and fidelity (data fit), and $\boldsymbol{I}(f,\psi_{w_{\psi}}(f))$ is the mutual information between the flow $f$ and its representation (code) $\psi_{w_{\psi}}(f)$. When $\beta$ is large, the encoder is lossy, thus preventing $Q_{w_{\varphi},w_{\psi}}(f|I)$ from being uninformative.\footnote{the decoder $\varphi$ imposes no architectural bottleneck due to skip connections.}

\subsection{Training a CPN}
\label{sect-training}

While the first term in Eq. \eqref{eq:final-optimization-CPN} can simply be the empirical autoencoding loss, the second term can be realized in many ways, {\em e.g.}, an $\ell^2$ or $\ell^1$ penalty on the parameters $w_{\psi}$. Here we directly increase the bottleneck $\beta$ by decreasing the coding length $\ell_{\psi}$ of $\psi$. Hence the training procedure of the proposed CPN can be summarized as follows:

\begin{enumerate}
  \item Initialize the coding length of the encoder $\ell_{\psi}$ with a large number ($\beta=0$).
  \item Train the encoder-decoder $\psi$, $\varphi$ jointly by minimizing $ e = \dfrac{1}{n}\sum_{j=1}^n \| \hat{f}_j - f_j \|^2$ until convergence. The error at convergence is denoted as $e^*$.
  \item If $e^* > \lambda$, training done.\footnote{in our experiments, $\lambda=0.5$.} \\
  Otherwise, decrease $\ell_{\psi}$, (increase $\beta$), and goto step 2.
\end{enumerate}
It would be time consuming to train for every single coding length $\ell_{\psi}$. We only iteratively train for the integer powers, $2^{k}, k\leq 10$.

\textbf{Inference}: suppose the optimal parameters obtained from the training procedure are $w_{\psi}^*$, $w_{\varphi}^*$, then for any given pair $(f, I)$, we can use $Q_{w_{\varphi}^*,w_{\psi}^*}(f|I)$ as the conditional prior up to a constant. In the next section we add a data discrepancy term to the (log) prior to obtain an energy functional for learning direct mapping from images to optical flows.

\subsection{Semi-unsupervised learning optical flow}

Unlike a generative model such as a variational autoencoder \cite{kingma2013auto}, where sampling is required in order to evaluate the probability of a given observation, here $(f, I)$ is directly mapped to a scalar using Eq. \eqref{eq:proposed-conditional-prior}, thus differentiable w.r.t $f$, and suitable for training a new network to predict optical flow given images $I_1,I_2$, by minimizing the following compound loss:
\begin{multline}
E(f|I_1,I_2) = \int_{\Omega \setminus O} \rho( I_1(x) - I_2(x+f(x)) )dx - \alpha \log [ Q_{w_{\varphi}^*,w_{\psi}^*}(f|I_1) ]\\
= \int_{\Omega \setminus O} \rho( I_1(x) - I_2(x+f(x)) )dx + \alpha \| \varphi^*(I_1, \psi^*(f)) - f \|^2
\label{eq:CPNFlow}
\end{multline}
with $\alpha>0$, $Q_{w_{\varphi}^*,w_{\psi}^*}$ our learned conditional prior, and $\rho(x) = (x^2+0.001^2)^{\eta}$ the generalized Charbonnier penalty function \cite{bruhn2005towards}. Note that the integration in the data term is on the co-visible area, i.e. the image domain $\Omega$ minus the occluded area $O$, which can be set to empty for simplicity or modeled using the forward-backward consistency as done in \cite{Meister:2018:UUL} with a penalty on $O$ to prevent trivial solutions. In the following section, we describe our implementation and report results and comparisons on several benchmarks.

\section{Experiments}

\subsection{Network details}

\textbf{CPN}: we adapt the FlowNetS network structure proposed in \cite{dosovitskiy2015flownet} to be the decoder $\varphi$, and the contraction part of FlowNetS to be the encoder $\psi$ in our CPN respectively. Both parts are shrunk versions of the original FlowNetS with a factor of $1\slash4$; altogether our CPN has $2.8$M parameters, which is an order of magnitude less than the $38$M parameters in FlowNetS. As we mentioned before, the bottleneck in Eq. \eqref{eq:final-optimization-CPN} is controlled by the coding length $\ell_{\psi}$ of the encoder $\psi$, here we make the definition of $\ell_{\psi}$ explicit, which is the number of the convolutional kernels in the last layer of the encoder. In our experiments, $\ell_{\psi}=128$ always satisfies the stopping criterion described in Sect. \ref{sect-training}, which ends up with a reduction rate $0.015$ in the dimension of the flow $f$.

\noindent \textbf{CPNFlow}: we term our flow prediction network CPNFlow. The network used on all benchmarks for comparison is the original FlowNetS with no modifications, letting us focus on the effects of different loss terms. The total number of parameters is $38$M. FlowNetS is the most basic network structure for learning optical flow \cite{dosovitskiy2015flownet}, {\em i.e.}, only convolutional layers with striding for dimension reduction, however, when trained with loss Eq. \eqref{eq:CPNFlow} that contains the learned conditional prior (CPN), it achieves better performance than the more complex network structure FlowNetC \cite{dosovitskiy2015flownet}, or even stack of FlowNetS and FlowNetC. Please refer to Sect. \ref{sect-benchmarks} for details and quantitative comparisons.

\subsection{Datasets for training}

\textbf{Flying Chairs} is a synthesized dataset proposed in \cite{dosovitskiy2015flownet}, by superimposing images of chairs on background images from Flickr. Randomly sampled 2-D affine transformations are applied to both chairs and background images. Thus there are independently moving objects together with background motion. The whole dataset contains about 22k $512 \times 384$ image pairs with ground truth flows.

\noindent \textbf{MPI-Sintel} \cite{butler2012naturalistic} is collected from an animation that made to be realistic. It contains scenes with natural illumination, objects moving fast, and articulated motion. Final and clean versions of the dataset are provided. The final version contains motion blur and fog effects. The training set contains only $1,041$ pairs of images, much smaller compared to Flying Chairs.

\noindent \textbf{KITTI} 2012 \cite{geiger2012we} and 2015 \cite{Menze2015CVPR} are the largest real-world datasets containing ground truth optical flows collected in a driving scenario. The ground truth flows are obtained from simultaneously recorded video and 3-D laser scans, together with some manual corrections. Even though the multi-view extended version contains roughly 15k image pairs, ground truth flows exist for only 394 pairs of image, which makes fully supervised training of optical flow prediction from scratch under this scenario infeasible. However, it provides a base for unsupervised learning of optical flow, and a stage to show the benefit of semi-unsupervised optical flow learning, that utilizes both the conditional prior (CPN) learned from the synthetic dataset, and the virtually unlimited amount of real world videos.

\subsection{Training details}

We use Adam \cite{kingma2014adam} as the optimizer with its default parameters in all our experiments. We train our conditional prior network (CPN) using Flying Chairs dataset due to its large amount of synthesized ground truth flows. The initial learning rate is 1.0e-4, and is halved every 100k steps until the maximum 600k training steps. The batch size is 8, and the autoencoding loss after training is around $0.6$.

There are two versions of our CPNFlow, i.e. CPNFlow-C and CPNFlow-K. Both employ the FlowNetS structure, and they differ in the training set on which Eq. \eqref{eq:CPNFlow} is minimized. CPNFlow-C is trained on Flying Chairs dataset, similarly CPNFlow-K is trained on KITTI dataset with the multi-view extension. The consideration here is: when trained on Flying Chairs dataset, the conditional prior network (CPN) is supposed to only capture the statistics of the affine transformations (a) CPNFlow-C is to test whether our learned prior works properly or not. If it works, (b) CPNFlow-K tests how the learned prior generalizes to real world scenarios. Both CPNFlow-C and CPNFlow-K have the same training schedule with the initial learning rate 1.0e-4, which is halved every 100k steps until the maximum 400k steps.\footnote{$\alpha=0.1, \eta=0.25$ for CPNFlow-C, and $\alpha=0.045, \eta=0.38$ for CPNFlow-K.} Note that in \cite{ren2017unsupervised}, layer-wise loss adjustment is used during training to simulate coarse-to-fine estimation, however, we will not adopt this training technique to avoid repeatedly interrupting the training process. In a similar spirit, we will not do network stacking as in \cite{Meister:2018:UUL,ilg2017flownet}, which increases both the training complexity and the network size.

In terms of data augmentation, we apply the same augmentation method as in \cite{dosovitskiy2015flownet} whenever our network is trained on Flying Chairs dataset with a cropping of 384x448. When trained on KITTI, resized to 384x512, only vertical flipping, horizontal flipping and image order switching are applied. The batch size used for training on Flying Chairs is 8 and on KITTI is 4.

\subsection{Benchmark results}
\label{sect-benchmarks}

\begin{table}[!h]
  \centering
  \begin{tabular}{P{0.3cm}|P{2.8cm}|P{1.0cm}|P{2.0cm}|P{2.0cm}|P{2.0cm}|P{2.0cm}}
    \hline
     &       & Chairs & Sintel Train & Sintel Test & KITTI Train & KITTI Test \\
     & Methods & test & clean\ \ \ final & clean\ \ \ final & 2012\ \ \ 2015 & 2012\ \ \ 2015\\
    \hline
    \multirow{4}{*}{\rotatebox[origin=c]{90}{
    \parbox[c]{1.2cm}{\centering Sup}}}
      & FlowNetS \cite{dosovitskiy2015flownet} & 2.71 & 4.50\ \ \ \ 5.45 & 7.42\ \ \ \ 8.43 & 8.26\ \ \ \ \ ----- & \textbf{9.1}\ \ \ \ \ ----- \\
      & FlowNetC \cite{dosovitskiy2015flownet} & \textbf{2.19} & 4.31\ \ \ \ 5.87 & 7.28\ \ \ \ 8.81 & 9.35\ \ \ \ \ ----- & -----\ \ \ \ \ ----- \\
      & SPyNet \cite{ranjan2017optical} & 2.63 & 4.12\ \ \ \ 5.57 & 6.69\ \ \ \ 8.43 & 9.12\ \ \ \ \ ----- & 10.1\ \ \ \ ----- \\
      & FlowNet2 \cite{ilg2017flownet} & ----- & \textbf{2.02}\ \ \ \textbf{3.14} & \textbf{3.96}\ \ \ \textbf{6.02} & \textbf{4.09}\ \ \ \textbf{10.06} & -----\ \ \ \ \ ----- \\
    \hline
    \multirow{4}{*}{\rotatebox[origin=c]{90}{
    \parbox[c]{1.2cm}{\centering Var}}}
      & Classic-NL \cite{sun2010secrets} & ----- & 6.03\ \ \ \ 7.99 & 7.96\ \ \ \ 9.15 & -----\ \ \ \ \ ----- & 16.4\ \ \ \ ----- \\
      & LDOF \cite{brox2011large} & \textbf{3.47} & \textbf{4.29}\ \ \ 6.42 & \textbf{7.56}\ \ \ \ \textbf{9.12} & 13.7\ \ \ \ ----- & 12.4\ \ \ \ ----- \\
      & HornSchunck \cite{sun2014quantitative}  & ----- & 7.23\ \ \ \ 8.38 & 8.73\ \ \ \ 9.61 & ----- \ \ \ \ ----- & \textbf{11.7}\ \ \textbf{41.8}\% \\
      & DIS-Fast \cite{kroeger2016fast}  & ----- & 5.61\ \ \ \textbf{6.31} & 9.35\ \ \ 10.13 & \textbf{11.01}\ \ \textbf{21.2} & 14.4\ \ \ \ ----- \\
    \hline
    \multirow{7}{*}{\rotatebox[origin=c]{90}{
    \parbox[c]{1.8cm}{\centering Unsup}}}
      & DSTFlow \cite{ren2017unsupervised} & 5.11 & 6.93\ \ \ \ 7.82 & 10.40\ \ \ 11.11 & 16.98\ \ \ 24.30 & -----\ \ \ \ \ ----- \\
      & DSTFlow-ft \cite{ren2017unsupervised} & 5.11 & (\textbf{6.16})\ (\textbf{6.81}) & 10.41\ \ \ 11.27 & 10.43\ \ \ 16.79 & 12.4\ \ \ \ \textbf{39}\% \\
      & BackToBasic \cite{jason2016back} & 5.30 & -----\ \ \ \ ----- & -----\ \ \ \ \ ----- & 11.30 \ \ \ \ ----- & \textbf{9.9}\ \ \ \ \ ----- \\
      & UnFlowC \cite{Meister:2018:UUL} & ----- & -----\ \ \ \ ----- & ----- \ \ \ \ ----- & 7.11\ \ \ \ 14.17 & -----\ \ \ \ \ ----- \\
      & UnFlowC-oc \cite{Meister:2018:UUL} & ----- & -----\ \ \ \ 8.64 & ----- \ \ \ \ ----- & 3.78\ \ \ \ \ 8.80 & -----\ \ \ \ \ ----- \\
      & UnFlowCSS-oc \cite{Meister:2018:UUL} & ----- & -----\ \ \ \ 7.91 & \textbf{9.37} \ \ \ 10.22 & \textbf{3.29}\ \ \ \ \textbf{8.10} & -----\ \ \ \ \ ----- \\
      & DenseNetF \cite{zhu2017densenet} & \textbf{4.73} & -----\ \ \ \ ----- & ----- \ \ \ \textbf{10.07} & -----\ \ \ \ \ ----- & 11.6\ \ \ \ \ ----- \\
    \hline
    \multirow{3}{*}{\rotatebox[origin=c]{90}{
    \parbox[c]{2cm}{\centering  }}}
      & CPNFlow-C & \textbf{3.81} & \textbf{4.87}\ \ \ \ \textbf{5.95} & \textbf{7.66}\ \ \ \ \textbf{8.58} & 7.33\ \ \ 14.61 & ----- \ \ \ \ \ ----- \\
      & CPNFlow-K & 4.37 & 6.46\ \ \ \ 7.12 & -----\ \ \ \ ----- & 3.76 \ \ \ 9.63 & 4.7 \ \ \ 30.8\% \\
      & CPNFlow-K-o & ----- & 7.01\ \ \ \ 7.52 & -----\ \ \ \ ----- & \textbf{3.11}\ \ \ \textbf{7.82} & \textbf{3.6}\ \ \ \textbf{30.4}\% \\
    \hline
  \end{tabular}
  \newline
  \caption{Quantitative evaluation and comparison to the state-of-the-art optical flow estimation methods coming from three different categories. Sup: Fully supervised, Var: Variational methods, and Unsup: Unsupervised learning methods. The performance measure is the end-point-error (EPE), except for the last column where percentage of erroneous pixels is used. The best performer in each category is highlighted in bold, and the number in parentheses is fine-tuned on the tested dataset. For more detailed comparisons on KITTI test sets, please refer to the online benchmark website: \url{http://www.cvlibs.net/datasets/kitti/eval_flow.php}.}
  \label{tab:quantative-comparison-all}
\end{table}

Tab. \ref{tab:quantative-comparison-all} summarizes our evaluation on all benchmarks mentioned above, together with quantitative comparisons to the state-of-the-art methods from different categories: Fully supervised, variational, and unsupervised learning methods. Since CPNFlow has the same network structure as FlowNetS, and both CPNFlow-C and FlowNetS are trained on Flying Chairs dataset, the comparison between CPNFlow-C and FlowNetS shows that even if CPNFlow-C is trained without knowing the correspondences between pairs of image and the ground truth flows, it can still achieve similar performance compared to the fully supervised ones on the synthetic dataset MPI-Sintel. When both are applied to KITTI, CPNFlow-C achieves $11.2\%$ and $21.6\%$ improvement over FlowNetS and FlowNetC respectively on KITTI 2012 Train, hence CPNFlow generalizes better to out of domain data.

One might notice that FlowNet2 \cite{ilg2017flownet} consistently achieves the highest score on MPI-Sintel and KITTI Train, however, it has a totally different network structure where several FlownetS \cite{dosovitskiy2015flownet} and FlowNetC \cite{dosovitskiy2015flownet} are stacked together, and it is trained in a sequential manner, and on additional datasets, e.g. FlyingThings3D \cite{mayer2016large} and a new dataset designed for small displacement \cite{ilg2017flownet}, thus not directly comparable to CPNFlow. However, when we simply apply the learned conditional prior to train our CPNFlow on KITTI using Eq. \eqref{eq:CPNFlow}, the final network CPNFlow-K surpasses FlowNet2 by $8\%$ on KITTI 2012 Train, yet the training procedure of CPNFlow is much simpler, and there is no need to switch between datasets nor between different modules of the network.

Since the emergence of unsupervised training of optical flow \cite{jason2016back}, there has not been a single method that beats the variational methods, as shown in Tab. \ref{tab:quantative-comparison-all}, even if both variational methods and unsupervised learning methods are minimizing the same type of loss function. One reason might be that when we implement the variational methods, we could apply some ``secret'' operations as mentioned in \cite{sun2010secrets}, e.g. median filtering, such that implicit regularization is triggered. Extra data term can also be added to bias the optimization, as in \cite{brox2011large}, sparse matches are used as a data term to deal with large displacements. However, when combined with our learned conditional prior, even the simplest data term would help unsupervisedly train a network that outperforms the state-of-the-art variational optical flow methods. As shown in Tab. \ref{tab:quantative-comparison-all} our CPNFlow consistently achieves similar or better performance than LDOF \cite{brox2011large}, especially on KITTI 2012 Train, the improvement is at least $40\%$.

Compared to unsupervised optical flow learning, the advantage of our learned conditional prior becomes obvious. Although DenseNetF \cite{zhu2017densenet} and UnFlowC \cite{Meister:2018:UUL} employ more powerful network structures than FlowNetS, their EPEs on MPI-Sintel Test are still 1.5 higher than our CPNFlow. Note that in \cite{Meister:2018:UUL}, several versions of result are reported, e.g. UnFlowC: trained with brightness data term and second order smoothness term, UnFlowC-oc: census transform based data term together with occlusion modeling and bidirectional flow consistency penalty, and UnFlowCSS-oc: a stack of one FlowNetC and two FlowNetS's sequentially trained using the same loss as in UnFlowC-oc. Our CPNFlow-K outperforms UnFlowC by $47\%$ on KITTI 2012 Train and $32\%$ on KITTI 2015 Train. When occlusion reasoning is effective in Eq. \eqref{eq:CPNFlow} as done in \cite{Meister:2018:UUL}, our CPNFlow-K-o outperforms UnFlowC-oc by $17.7\%$ on KITTI 2012 Train, $11.1\%$ on KITTI 2015 Train, and $12.9\%$ on Sintel Train Final, even without a more robust census transform based data term and flow consistency penalty, which demonstrate the effectiveness of our learned conditional prior across different data terms. Note that our CPNFlow-K-o even outperforms UnFlowCSS-oc, which is far more complex in training and network architecture.

\begin{figure}[!ht]
  \centering
  \includegraphics[width=0.85\textwidth]{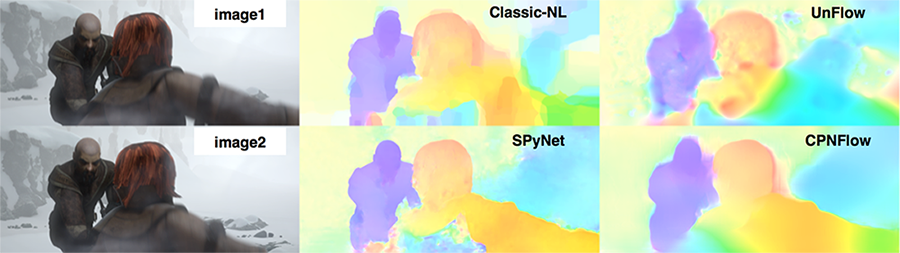}
  \includegraphics[width=0.85\textwidth]{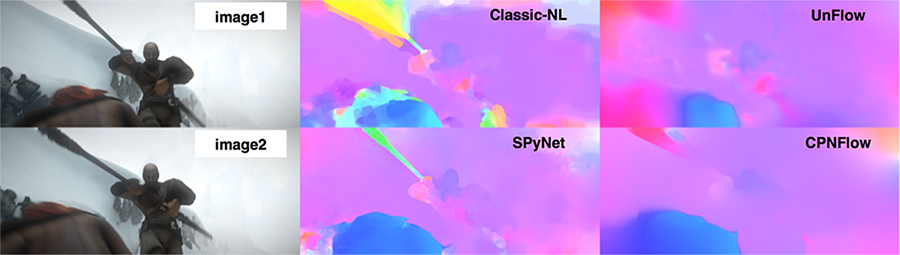}
  \includegraphics[width=0.85\textwidth]{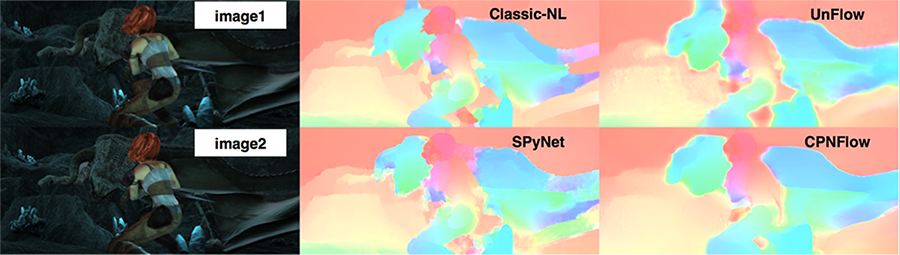}
  \includegraphics[width=0.85\textwidth]{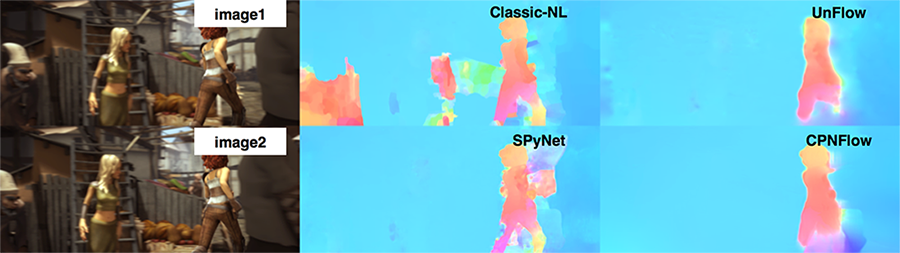}
  \includegraphics[width=0.85\textwidth]{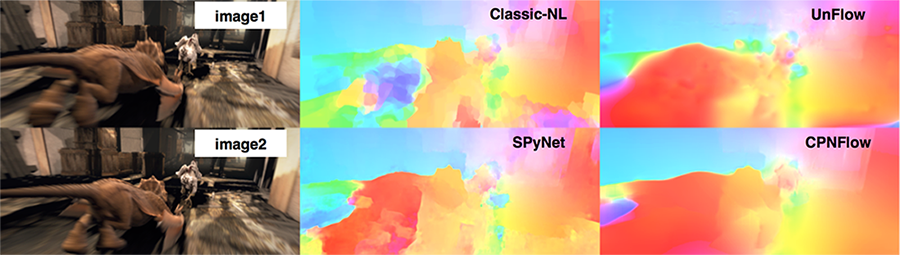}
  \caption{Visual comparison on MPI-Sintel. Variational: CLassic-NL \cite{sun2010secrets}, Supervised: SPyNet \cite{ranjan2017optical}, Unsupervised: UnFlowC \cite{Meister:2018:UUL} and our CPNFlow-C.}
  \label{fig:visuals-sintel}
\end{figure}

\begin{figure}[!ht]
  \centering
  \includegraphics[width=0.88\textwidth]{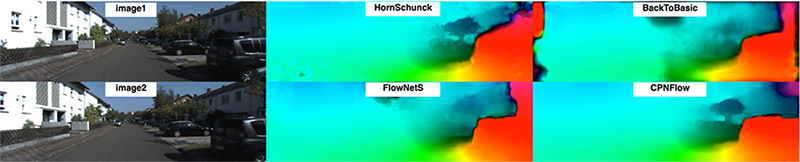}
  \includegraphics[width=0.88\textwidth]{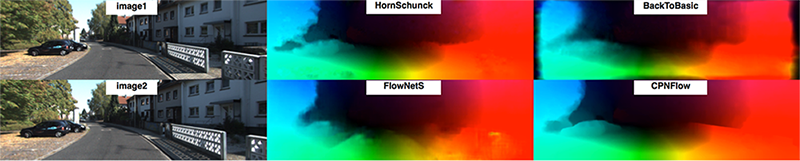}
  \includegraphics[width=0.88\textwidth]{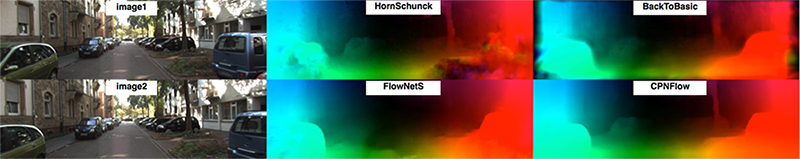}
  \includegraphics[width=0.88\textwidth]{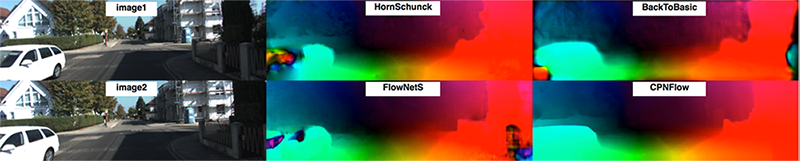}
  \includegraphics[width=0.88\textwidth]{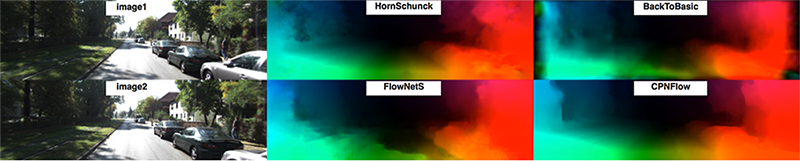}
  \caption{Visual comparison on KITTI 2012. Variational: HornSchunck \cite{sun2014quantitative}, Supervised: FlowNetS \cite{dosovitskiy2015flownet}, Unsupervised: BackToBasic \cite{jason2016back} and our CPNFlow-K.}
  \label{fig:visuals-kitti}
\end{figure}

\begin{figure}[!ht]
  \centering
  \includegraphics[width=0.88\textwidth]{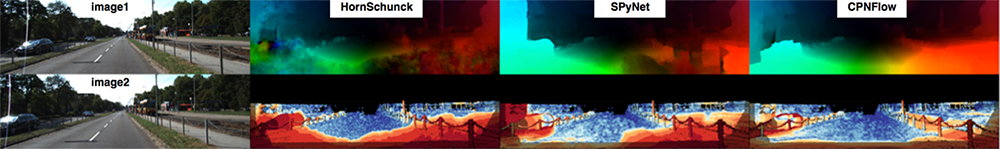}
  \includegraphics[width=0.88\textwidth]{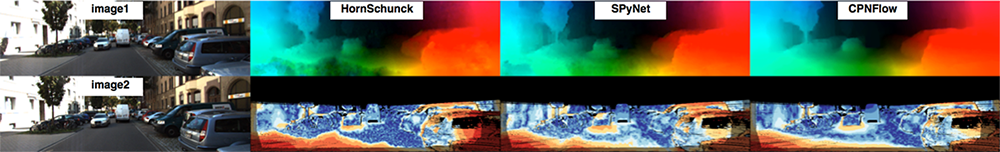}
  \includegraphics[width=0.88\textwidth]{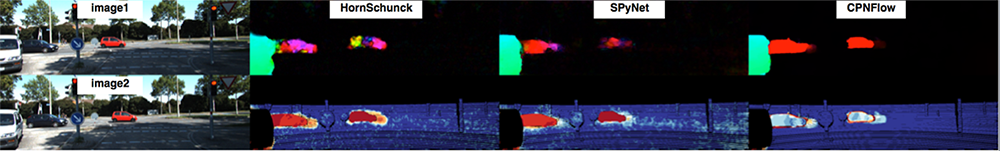}
  \includegraphics[width=0.88\textwidth]{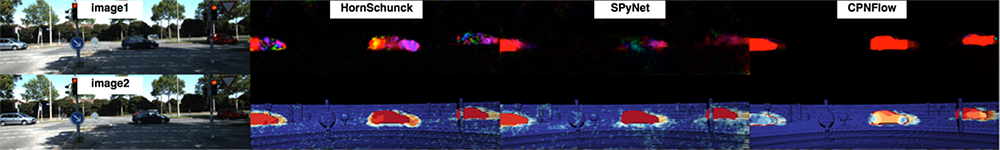}
  \includegraphics[width=0.88\textwidth]{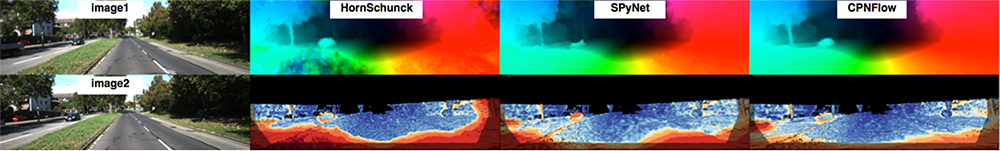}
  \includegraphics[width=0.88\textwidth]{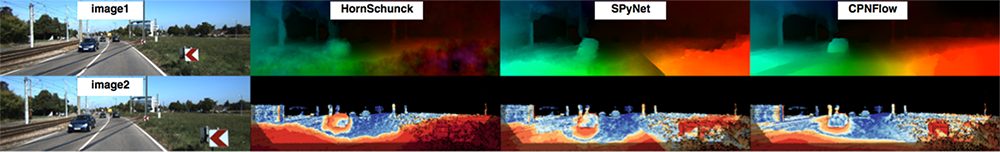}
  \includegraphics[width=0.88\textwidth]{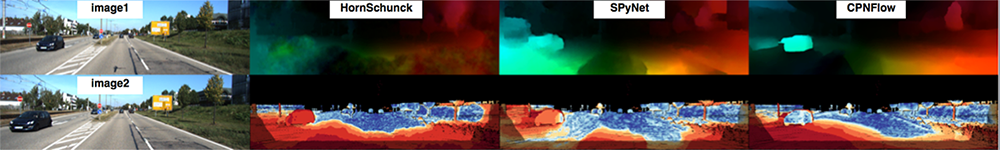}
  \caption{Visual comparison on KITTI 2015. Variational: HornSchunck \cite{sun2014quantitative}, Supervised: SPyNet \cite{ranjan2017optical} and our CPNFlow-K. The 2nd row in each pair shows the end-point-error of the estimated flow, red is high and blue is low.}
  \label{fig:visuals-kitti-2015}
\end{figure}

Fig. \ref{fig:visuals-sintel}, Fig. \ref{fig:visuals-kitti}, Fig. \ref{fig:visuals-kitti-2015} show the visual comparisons on MPI-Sintel, KITTI 2012 and KITTI 2015 respectively. Note that our CPNFlow is generally much smoother, and at the same time sharper at object boundaries, e.g. the girl in the 3rd, 4th rows and the dragon in the 5th row in Fig. \ref{fig:visuals-sintel}. This demonstrates that our conditional prior network (CPN) is capable of learning high level (semantic) regularities imposed by object entities. In Fig. \ref{fig:visuals-kitti}, we can also observe that discontinuities in the flow fields align well with object boundaries, for example the cars in all pairs. This, again, demonstrates that our learned conditional prior is able to generalize to different scenarios. The error of the estimated flows is also displayed in Fig. \ref{fig:visuals-kitti-2015}.

\section{Discussion and related work}

Generic priors capturing rudimentary statistics to regularize optical flow have been used for decades, starting with Horn \& Schunk's $\ell^2$ norm of the gradient, to $\ell^1$, Total Variation, etc. We seek to design or learn image-dependent priors that capture long-range correlation and semantics.

Image-dependent priors of the form Eq. \eqref{eq:classical-of} include \cite{krahenbuhl2012efficient,ranftl2014non,chen2016full,deriche1995optical,proesmans1994determination,brox2004high,xu2012motion}, whereas most recent methods learn optical flow end-to-end, without explicitly isolating the likelihood and prior terms, for instance \cite{dosovitskiy2015flownet,ilg2017flownet,ranjan2017optical} are the top performing on MPI-Sintel. Some methods even cast optical flow as dense or semi-dense feature matching \cite{bailer2015flow,liu2008sift,Yang_2015_CVPR,yang2017s2f} in order to deal with large displacements, while the regularity is merely imposed by forward-backward matching consistency (see references therein for a detailed review of related literature).

It would be tempting to use a GAN \cite{goodfellow2014generative} to learn the prior distribution of interest. A GAN can be thought of as a method to learn a map $g$ such that its push-forward $g_*$ maps two distributions, one known $\mu$, and one we can sample from, $p$, so $\hat g = \arg\min {\mathbb{KL}}(g_* \mu || p)$. It does so via an adversarial process such that a generative model $G$ will capture the data distribution $p_{data}$. If we sample from the generative model $G$, we will have samples that are equivalently sampled from $p_{data}$, in order to evaluate $p_{data}(x)$ of a sample $x$, we can not circumvent the sampling step, thus making the method unsuitable for our purpose where we want a differentiable scalar function.

Our work entails constructing an autoencoder of the flow, so it naturally relates to \cite{kingma2013auto}. Similarly, evaluating the probability of a test example is intractable, even if we can approximately evaluate the lower bound of the probability of a data point, which again can not be computed in closed form due to the expectation over the noise.

Optical flow learning algorithms typically rely on synthesized datasets, due to the extreme difficulty in obtaining ground truth flows for realistic videos. Recently, unsupervised optical flow learning methods have flourished, making use of vast amount of unlabeled videos. Although unsupervised optical flow learning methods are able to learn from unlimited amount of data, when compared to variational methods, their performance usually falls behind, even when a similar loss is employed. A phenomenon observed is that almost all unsupervised optical flow learning methods use the Horn-Schunck type surrogate losses. And there is debate on which feature to use for the data term, {\em e.g.}, the raw photometric value or the edge response, or on the prior/regularizer term, {\em e.g. }, penalizing the first order gradient of the flow or the second order, or on how to weight the prior term in a pixel-wise manner. Surrogate losses are getting more and more complex. Instead of focusing on the data term, we ask what should be the best form for the prior term. Our answer is that structural consistency between an image and the flow, as well as high-order statistics, such as semantic consistency, are important. We show that when combined with the raw photometric warping error, this kind of prior serves as a better regularizer than all the other hand-designed ones. We show its effectiveness on several contemporary optical flow benchmarks, also thanks to its ability to leverage existing limited supervised (synthetic) datasets and unlimited real world videos.

\section{Appendix}

\subsection{Equivalence between Eq. (4) and Eq. (5)}

Using definition of Kullback-Leibler divergence we have:

\begin{multline}
w_{\varphi}^*, w_{\psi}^* = \arg\min_{w_{\varphi}, w_{\psi}} {\mathbb E}_{I\sim P(I)} KL( P(f|I) \| Q_{w_{\varphi}, w_{\psi}}(f|I) )\\
= \arg\min_{w_{\varphi}, w_{\psi}} \int_I P(I) KL( P(f|I) \| Q_{w_{\varphi}, w_{\psi}}(f|I) )\\
= \arg\min_{w_{\varphi}, w_{\psi}} \int_I P(I) \int_f ( P(f|I) log\dfrac{P(f|I)}{Q_{w_{\varphi}, w_{\psi}}(f|I)} )df dI\\
= \arg\min_{w_{\varphi}, w_{\psi}} \int_I \int_f P(I)P(f|I)logP(f|I)df dI - \int_I \int_f P(I)P(f|I) log Q_{w_{\varphi}, w_{\psi}}(f|I) df dI\\
= \arg\max_{w_{\varphi}, w_{\psi}} \int_I \int_f P(f, I) log Q_{w_{\varphi}, w_{\psi}}(f|I) df dI\\
= \arg\max_{w_{\varphi}, w_{\psi}} - \int_I \int_f P(f, I) \| \varphi_{w_\varphi}(I, \psi_{w_\psi}(f)) - f \|^2 df dI \\
= \arg\min_{w_{\varphi}, w_{\psi}} \int_I \int_f P(f, I) \| \varphi_{w_\varphi}(I, \psi_{w_\psi}(f)) - f \|^2 df dI
\label{eq:autoencoding-loss}
\end{multline}

\subsection{Reasoning behind the CPN structure}

We are going to show our reasoning that leads us to the current CPN structure instead of the ordinary autoencoder which encodes both $f$ and $I$ in one branch as follows:

\begin{equation}
Q_{w_{\varphi},w_{\psi}}(f|I) = \exp \left( - \| \varphi \circ \psi(f, I) - f \|^2 \right)
\label{eq:temp-proposed-conditional-prior}
\end{equation}

where $\varphi$ is a decoder and $\psi$ is an encoder of both $f$ and $I$, parameterized by $w_{\varphi}, w_{\psi}$ respectively. The optimal parameters $w_{\varphi}^*, w_{\psi}^*$ are obtained by minimizing the average KL divergence between the proposed conditional $Q$ and $P(f|I)$:

\begin{equation}
w_{\varphi}^*, w_{\psi}^* = \arg\min_{w_{\varphi}, w_{\psi}} E_{I\sim P(I)} KL( P(f|I) \| Q_{w_{\varphi}, w_{\psi}}(f|I) )
\end{equation}

similarly to former subsection, we can show that the above optimization problem is equivalent to:

\begin{multline}
w_{\varphi}^*, w_{\psi}^* = \arg\max_{w_{\varphi}, w_{\psi}} \int_I \int_f P(f, I) \log [ Q_{w_{\varphi}, w_{\psi}}(f|I) ]df dI \\
= \arg\min_{w_{\varphi}, w_{\psi}} \int_I \int_f P(f, I) \| \varphi_{w_\varphi} \circ \psi_{w_\psi}(f, I) - f \|^2 df dI
\label{eq:temp-equivalent-optimization-formula}
\end{multline}

However, we are not done as $\psi$ is an encoder with limited capacity, thus $\psi$ is not one-to-one, which makes the following subset non-empty:

\begin{equation}
\boldsymbol{I}_{\psi, f} = \{ I | \psi_{w_\psi}(f, I) = \psi \}.
\label{eq:temp-set-I}
\end{equation}

We can rewrite the optimization problem Eq. \eqref{eq:temp-equivalent-optimization-formula} as:

\begin{multline}
w_{\varphi}^*, w_{\psi}^* = \arg\min_{w_{\varphi}, w_{\psi}} \int_f \int_{\psi} \int_{I \in \boldsymbol{I}_{\psi, f}} P(f, I) \| \varphi_{w_\varphi} \circ \psi_{w_\psi}(f, I) - f \|^2 dI d\psi df \\
= \arg\min_{w_{\varphi}, w_{\psi}} \int_f \int_{\psi} \| \varphi_{w_\varphi}(\psi) - f \|^2 \left( \int_{I \in \boldsymbol{I}_{\psi, f}} P(f, I)dI \right) d\psi df \\
= \arg\min_{w_{\varphi}, w_{\psi}} \int_f \int_{\psi} \| \varphi_{w_\varphi}(\psi) - f \|^2 P(f)P_{w_\psi}(\psi |f) d\psi df
\end{multline}

$P_{w_\psi}(\psi | f)$ is a probability measure induced by the encoder $\psi$. Thus, the original optimization problem is essentially minimizing the following quantity:

\begin{equation}
w_{\varphi}^*, w_{\psi}^* = KL\left( P(f)P_{w_\psi}(\psi | f) \| Q_{w_\varphi}(\psi, f) \right)
\label{eq:temp-essential-optimization-formula}
\end{equation}

During the optimization process, the encoder is trying to push $P(f)P_{w_\psi}(\psi | f)$ towards $Q_{w_\varphi}(\psi, f)$ and the decoder is pushing from the other side. After optimization:

\begin{equation}
Q_{w_{\varphi},w_{\psi}}(f|I) = \exp \left( - \| \varphi \circ \psi(f, I) - f \|^2 \right) \propto P(f)P_{w_\psi}(\psi(f,I) | f)
\end{equation}

which is not $P(f|I)$ nor $P(f,I)$! In order to let $Q_{w_{\varphi},w_{\psi}}(f|I)$ approximate $P(f|I)$, the condition $P_{w_\psi}(\psi|f) = P(I|f)$ should be true. And this is satisfied when $\psi$ imposes no compression on $I$. i.e. $\psi: (f, I) \rightarrow (\psi(f), I)$, which enforces Eq. \eqref{eq:temp-set-I} to be a singleton, and now we have the proposed CPN structure.

\subsection{Relation to Information Bottleneck(IB)}

In \cite{achille2017emergence}, the IB Lagrangian is proposed to achieve the sufficiency and the minimality of the learned representation. In our CPN training loss (Eq. (6) in the main paper), the first term is equivalent to the autoencoding loss Eq. \eqref{eq:autoencoding-loss}, minimizing which is equivalent to maximizing the following conditional mutual information:

\begin{equation}
\boldsymbol{I}( f, \psi_{w_\psi}(f) | I )
\end{equation}

which measures the conditional sufficiency, together with the second term that measures the minimality of the representation $\psi_{w_{\psi}}(f)$, our CPN training loss can be viewed as  minimizing:

\begin{equation}
E(w_{\varphi}, w_{\psi}; D) = - \boldsymbol{I}( f, \psi_{w_\psi}(f) | I ) + 
\beta \boldsymbol{I}(f, \psi_{w_{\psi}}(f))
\end{equation}

which is in line with the principles proposed in \cite{achille2017emergence}.

\bibliographystyle{splncs04}
\bibliography{egbib}

\begin{thebibliography}{10}
\providecommand{\url}[1]{\texttt{#1}}
\providecommand{\urlprefix}{URL }
\providecommand{\doi}[1]{https://doi.org/#1}

\bibitem{achille2017emergence}
Achille, A., Soatto, S.: Emergence of invariance and disentangling in deep
  representations. Journal of Machine Learning Research (JMLR), in press. Also
  in Proc. of the ICML Workshop on Principled Approaches to Deep Learning;
  ArXiv 1706.01350  (May 30, 2017)

\bibitem{ahmadi2016unsupervised}
Ahmadi, A., Patras, I.: Unsupervised convolutional neural networks for motion
  estimation. In: Image Processing (ICIP), 2016 IEEE International Conference
  on. pp. 1629--1633. IEEE (2016)

\bibitem{bailer2015flow}
Bailer, C., Taetz, B., Stricker, D.: Flow fields: Dense correspondence fields
  for highly accurate large displacement optical flow estimation. In:
  Proceedings of the IEEE international conference on computer vision. pp.
  4015--4023 (2015)

\bibitem{baker2011database}
Baker, S., Scharstein, D., Lewis, J., Roth, S., Black, M.J., Szeliski, R.: A
  database and evaluation methodology for optical flow. International Journal
  of Computer Vision  \textbf{92}(1),  1--31 (2011)

\bibitem{black1993framework}
Black, M.J., Anandan, P.: A framework for the robust estimation of optical
  flow. In: Computer Vision, 1993. Proceedings., Fourth International
  Conference on. pp. 231--236. IEEE (1993)

\bibitem{brox2004high}
Brox, T., Bruhn, A., Papenberg, N., Weickert, J.: High accuracy optical flow
  estimation based on a theory for warping. In: European conference on computer
  vision. pp. 25--36. Springer (2004)

\bibitem{brox2011large}
Brox, T., Malik, J.: Large displacement optical flow: descriptor matching in
  variational motion estimation. IEEE transactions on pattern analysis and
  machine intelligence  \textbf{33}(3),  500--513 (2011)

\bibitem{bruhn2005towards}
Bruhn, A., Weickert, J.: Towards ultimate motion estimation: Combining highest
  accuracy with real-time performance. In: Computer Vision, 2005. ICCV 2005.
  Tenth IEEE International Conference on. vol.~1, pp. 749--755. IEEE (2005)

\bibitem{bruhn2005lucas}
Bruhn, A., Weickert, J., Schn{\"o}rr, C.: Lucas/kanade meets horn/schunck:
  Combining local and global optic flow methods. International journal of
  computer vision  \textbf{61}(3),  211--231 (2005)

\bibitem{butler2012naturalistic}
Butler, D.J., Wulff, J., Stanley, G.B., Black, M.J.: A naturalistic open source
  movie for optical flow evaluation. In: European Conference on Computer
  Vision. pp. 611--625. Springer (2012)

\bibitem{chaudhari2017stochastic}
Chaudhari, P., Soatto, S.: Stochastic gradient descent performs variational
  inference, converges to limit cycles for deep networks. arXiv preprint
  arXiv:1710.11029  (2017)

\bibitem{chen2016full}
Chen, Q., Koltun, V.: Full flow: Optical flow estimation by global optimization
  over regular grids. In: Proceedings of the IEEE Conference on Computer Vision
  and Pattern Recognition. pp. 4706--4714 (2016)

\bibitem{cohen2016inductive}
Cohen, N., Shashua, A.: Inductive bias of deep convolutional networks through
  pooling geometry. arXiv preprint arXiv:1605.06743  (2016)

\bibitem{deriche1995optical}
Deriche, R., Kornprobst, P., Aubert, G.: Optical-flow estimation while
  preserving its discontinuities: A variational approach. In: Asian Conference
  on Computer Vision. pp. 69--80. Springer (1995)

\bibitem{dosovitskiy2015flownet}
Dosovitskiy, A., Fischer, P., Ilg, E., Hausser, P., Hazirbas, C., Golkov, V.,
  van~der Smagt, P., Cremers, D., Brox, T.: Flownet: Learning optical flow with
  convolutional networks. In: Proceedings of the IEEE International Conference
  on Computer Vision. pp. 2758--2766 (2015)

\bibitem{geiger2012we}
Geiger, A., Lenz, P., Urtasun, R.: Are we ready for autonomous driving? the
  kitti vision benchmark suite. In: Computer Vision and Pattern Recognition
  (CVPR), 2012 IEEE Conference on. pp. 3354--3361. IEEE (2012)

\bibitem{goodfellow2014generative}
Goodfellow, I., Pouget-Abadie, J., Mirza, M., Xu, B., Warde-Farley, D., Ozair,
  S., Courville, A., Bengio, Y.: Generative adversarial nets. In: Advances in
  neural information processing systems. pp. 2672--2680 (2014)

\bibitem{horn1981determining}
Horn, B.K., Schunck, B.G.: Determining optical flow. Artificial intelligence
  \textbf{17}(1-3),  185--203 (1981)

\bibitem{ilg2017flownet}
Ilg, E., Mayer, N., Saikia, T., Keuper, M., Dosovitskiy, A., Brox, T.: Flownet
  2.0: Evolution of optical flow estimation with deep networks. In: Proceedings
  of the IEEE Conference on Computer Vision and Pattern Recognition. pp.
  2462--2470 (2017)

\bibitem{jason2016back}
Jason, J.Y., Harley, A.W., Derpanis, K.G.: Back to basics: Unsupervised
  learning of optical flow via brightness constancy and motion smoothness. In:
  European Conference on Computer Vision. pp. 3--10. Springer (2016)

\bibitem{kingma2014adam}
Kingma, D.P., Ba, J.: Adam: A method for stochastic optimization. arXiv
  preprint arXiv:1412.6980  (2014)

\bibitem{kingma2013auto}
Kingma, D.P., Welling, M.: Auto-encoding variational bayes. arXiv preprint
  arXiv:1312.6114  (2013)

\bibitem{krahenbuhl2012efficient}
Kr{\"a}henb{\"u}hl, P., Koltun, V.: Efficient nonlocal regularization for
  optical flow. In: European Conference on Computer Vision. pp. 356--369.
  Springer (2012)

\bibitem{kroeger2016fast}
Kroeger, T., Timofte, R., Dai, D., Van~Gool, L.: Fast optical flow using dense
  inverse search. In: European Conference on Computer Vision. pp. 471--488.
  Springer (2016)

\bibitem{liu2008sift}
Liu, C., Yuen, J., Torralba, A., Sivic, J., Freeman, W.T.: Sift flow: Dense
  correspondence across different scenes. In: European conference on computer
  vision. pp. 28--42. Springer (2008)

\bibitem{mayer2016large}
Mayer, N., Ilg, E., Hausser, P., Fischer, P., Cremers, D., Dosovitskiy, A.,
  Brox, T.: A large dataset to train convolutional networks for disparity,
  optical flow, and scene flow estimation. In: Proceedings of the IEEE
  Conference on Computer Vision and Pattern Recognition. pp. 4040--4048 (2016)

\bibitem{Meister:2018:UUL}
Meister, S., Hur, J., Roth, S.: {UnFlow}: Unsupervised learning of optical flow
  with a bidirectional census loss. In: AAAI. New Orleans, Louisiana (Feb 2018)

\bibitem{Menze2015CVPR}
Menze, M., Geiger, A.: Object scene flow for autonomous vehicles. In:
  Conference on Computer Vision and Pattern Recognition (CVPR) (2015)

\bibitem{papenberg2006highly}
Papenberg, N., Bruhn, A., Brox, T., Didas, S., Weickert, J.: Highly accurate
  optic flow computation with theoretically justified warping. International
  Journal of Computer Vision  \textbf{67}(2),  141--158 (2006)

\bibitem{proesmans1994determination}
Proesmans, M., Van~Gool, L., Pauwels, E., Oosterlinck, A.: Determination of
  optical flow and its discontinuities using non-linear diffusion. In: European
  Conference on Computer Vision. pp. 294--304. Springer (1994)

\bibitem{ranftl2014non}
Ranftl, R., Bredies, K., Pock, T.: Non-local total generalized variation for
  optical flow estimation. In: European Conference on Computer Vision. pp.
  439--454. Springer (2014)

\bibitem{ranjan2017optical}
Ranjan, A., Black, M.J.: Optical flow estimation using a spatial pyramid
  network. In: Proceedings of the IEEE Conference on Computer Vision and
  Pattern Recognition. pp. 4161--4170 (2017)

\bibitem{ren2017unsupervised}
Ren, Z., Yan, J., Ni, B., Liu, B., Yang, X., Zha, H.: Unsupervised deep
  learning for optical flow estimation. In: Thirty-First AAAI Conference on
  Artificial Intelligence (2017)

\bibitem{sun2010secrets}
Sun, D., Roth, S., Black, M.J.: Secrets of optical flow estimation and their
  principles. In: Computer Vision and Pattern Recognition (CVPR), 2010 IEEE
  Conference on. pp. 2432--2439. IEEE (2010)

\bibitem{sun2014quantitative}
Sun, D., Roth, S., Black, M.J.: A quantitative analysis of current practices in
  optical flow estimation and the principles behind them. International Journal
  of Computer Vision  \textbf{106}(2),  115--137 (2014)

\bibitem{sutton2012introduction}
Sutton, C., McCallum, A., et~al.: An introduction to conditional random fields.
  Foundations and Trends{\textregistered} in Machine Learning  \textbf{4}(4),
  267--373 (2012)

\bibitem{xu2012motion}
Xu, L., Jia, J., Matsushita, Y.: Motion detail preserving optical flow
  estimation. IEEE Transactions on Pattern Analysis and Machine Intelligence
  \textbf{34}(9),  1744--1757 (2012)

\bibitem{Yang_2015_CVPR}
Yang, Y., Lu, Z., Sundaramoorthi, G., et~al.: Coarse-to-fine region selection
  and matching. In: 2015 IEEE Conference on Computer Vision and Pattern
  Recognition (CVPR). pp. 5051--5059. IEEE (2015)

\bibitem{yang2017s2f}
Yang, Y., Soatto, S.: S2f: Slow-to-fast interpolator flow. In: 2017 IEEE
  Conference on Computer Vision and Pattern Recognition (CVPR). pp. 3767--3776.
  IEEE (2017)

\bibitem{zhu2017densenet}
Zhu, Y., Newsam, S.: Densenet for dense flow. arXiv preprint arXiv:1707.06316
  (2017)

\end{thebibliography}
\end{document}